\documentclass[final]{cvpr}

\usepackage{times}
\usepackage{epsfig}
\usepackage{graphicx}
\usepackage{amsmath}
\usepackage{amssymb}
\usepackage{booktabs}
\usepackage{enumitem}
\usepackage{physics}
\usepackage{multirow}

\makeatletter
\@namedef{ver@everyshi.sty}{}
\makeatother
\usepackage{tikz}
\usepackage{pgfplots}

\usepackage{hhline}
\usepackage[page]{appendix}
\DeclareFixedFont{\ttb}{T1}{txtt}{bx}{n}{12} % for bold
\DeclareFixedFont{\ttm}{T1}{txtt}{m}{n}{12}  % for normal

\newcommand{\pmm}[1]{p_{m}}
\newcommand{\method}[1]{USI}

\usepackage{subcaption}
\usepackage{makecell} 
\usepackage{graphicx}

\usepackage{amsmath,amsfonts,bm}

\def\eqref#1{equation~\ref{#1}}
\def\1{\bm{1}}

\DeclareMathAlphabet{\mathsfit}{\encodingdefault}{\sfdefault}{m}{sl}
\SetMathAlphabet{\mathsfit}{bold}{\encodingdefault}{\sfdefault}{bx}{n}

\usepackage{float}
\restylefloat{table}
\usepackage[pagebackref=true,breaklinks=true,colorlinks,bookmarks=false]{hyperref}

\begin{document}

%%%%%%%%% TITLE
% \title{Solving ImageNet: a Unified Training Scheme for Obtaining Top Results on Any Backbone}
\title{Solving ImageNet: a Unified Scheme for Training any Backbone to Top Results}

\author{Tal Ridnik, Hussam Lawen, Emanuel Ben-Baruch, Asaf Noy\\
DAMO Academy, Alibaba Group\\
{\tt\small tal.ridnik@alibaba-inc.com}
% \and
}

\maketitle

\begin{abstract}
\label{sec:abstract}
ImageNet serves as the primary dataset for evaluating the quality of computer-vision models. The common practice today is training each architecture with a tailor-made scheme, designed and tuned by an expert.
In this paper, we present a unified scheme for training any backbone on ImageNet. The scheme, named \method{} (Unified Scheme for ImageNet), is based on knowledge distillation and modern tricks.  It requires no adjustments or hyper-parameters tuning between different models, and is efficient in terms of training times.
We test \method{} on a wide variety of architectures, including CNNs, Transformers, Mobile-oriented and MLP-only. On all models tested, \method{} outperforms previous state-of-the-art results. Hence, we are able to transform training on ImageNet from an expert-oriented task to an automatic seamless routine.
Since \method{} accepts any backbone and trains it to top results, it also enables to perform methodical comparisons, and identify the most efficient backbones along the speed-accuracy Pareto curve.
\\
Implementation is available at: \url{https://github.com/Alibaba-MIIL/Solving\_ImageNet}
\end{abstract}

\section{Introduction}

\begin{figure*}[t!]
\centering
\begin{subfigure}[a]{.48\textwidth}
  \centering
  \includegraphics[scale=0.23]{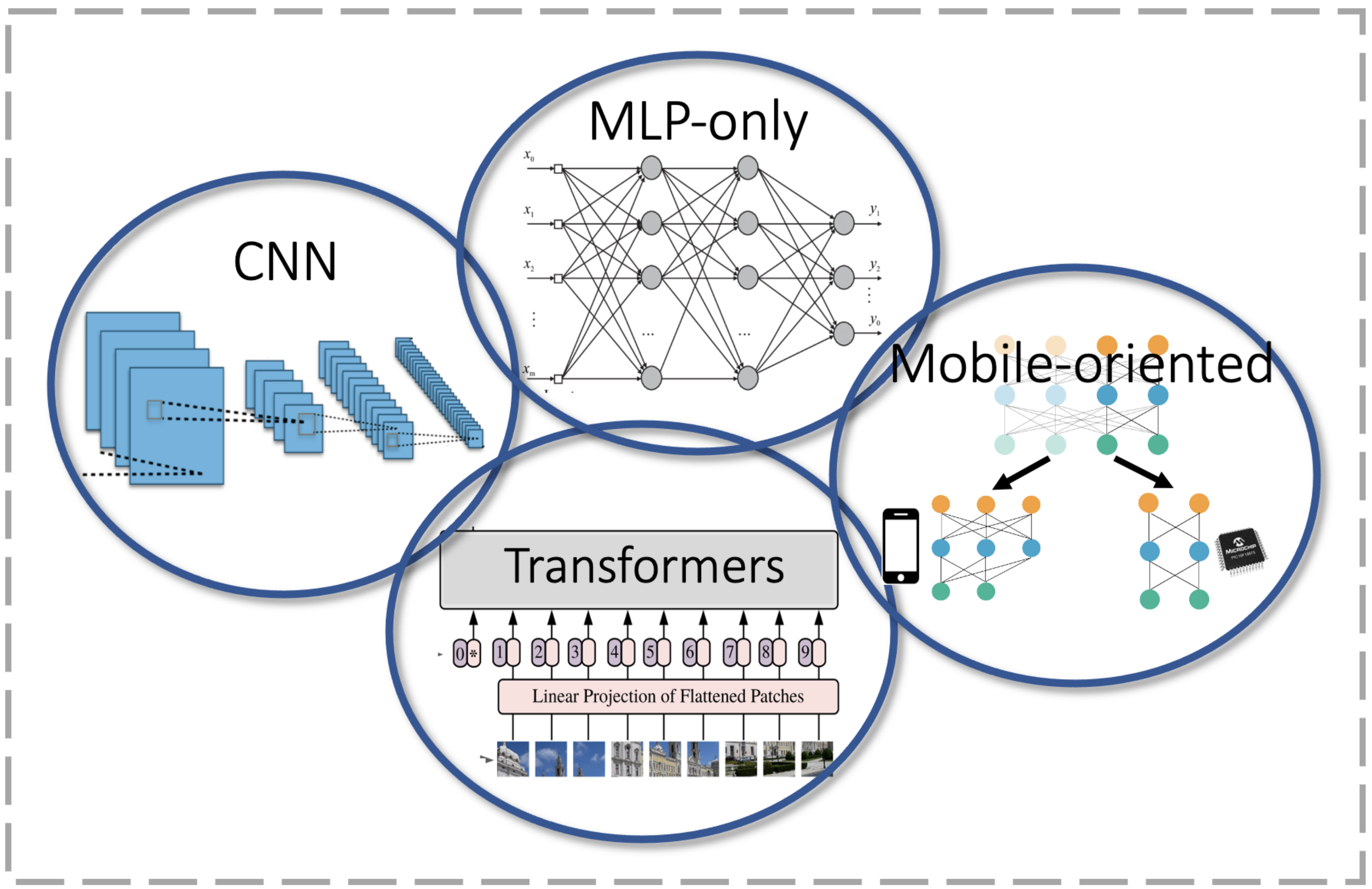}
  \caption{Various architecture families}
\end{subfigure}%
\begin{subfigure}[a]{.44\textwidth }
  \centering
  \includegraphics[scale=0.35]{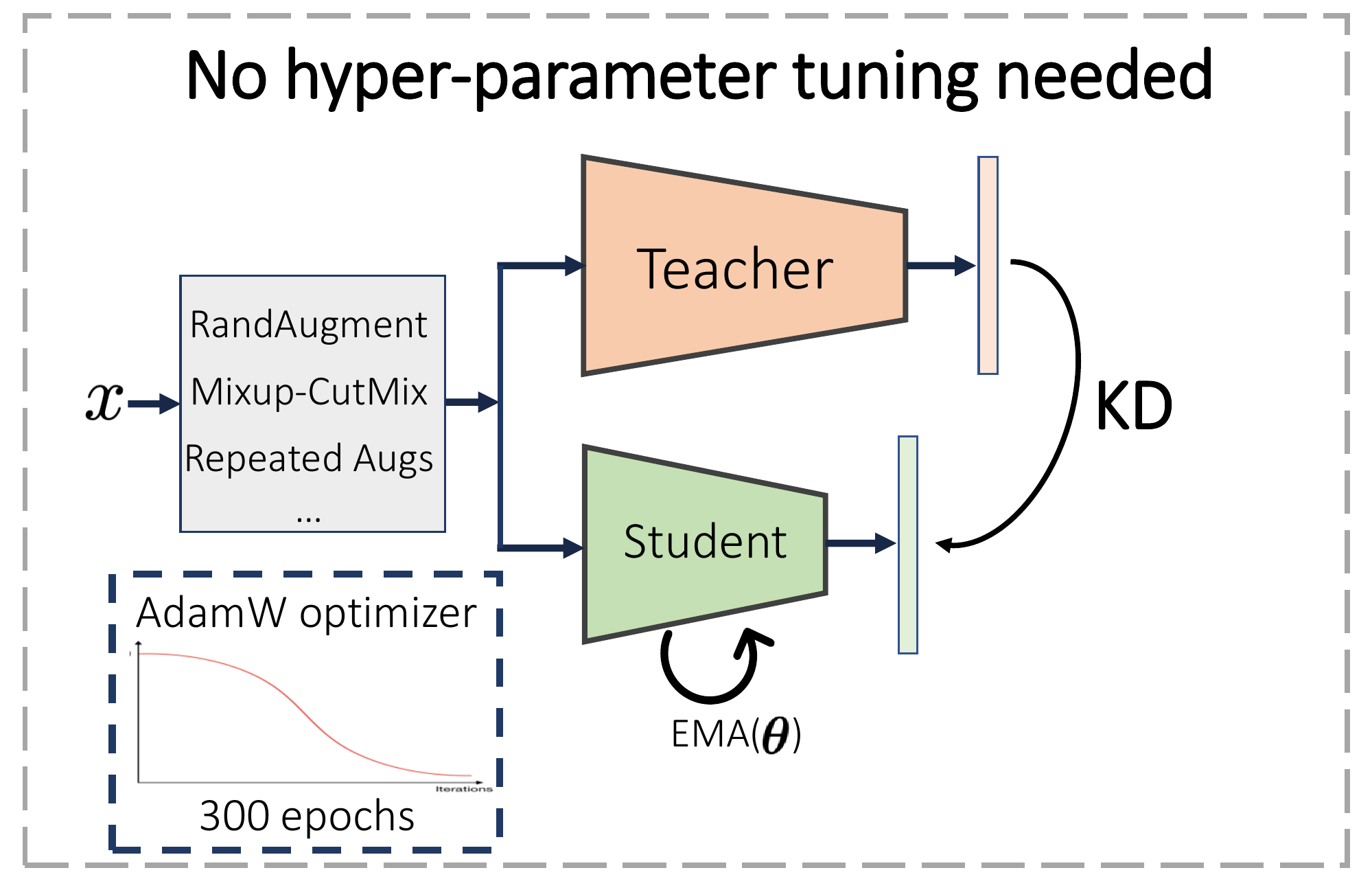}
  \caption{\method{} training scheme}
\end{subfigure}
\begin{subfigure}[a]{.99\textwidth }
  \centering
  \includegraphics[scale=0.55]{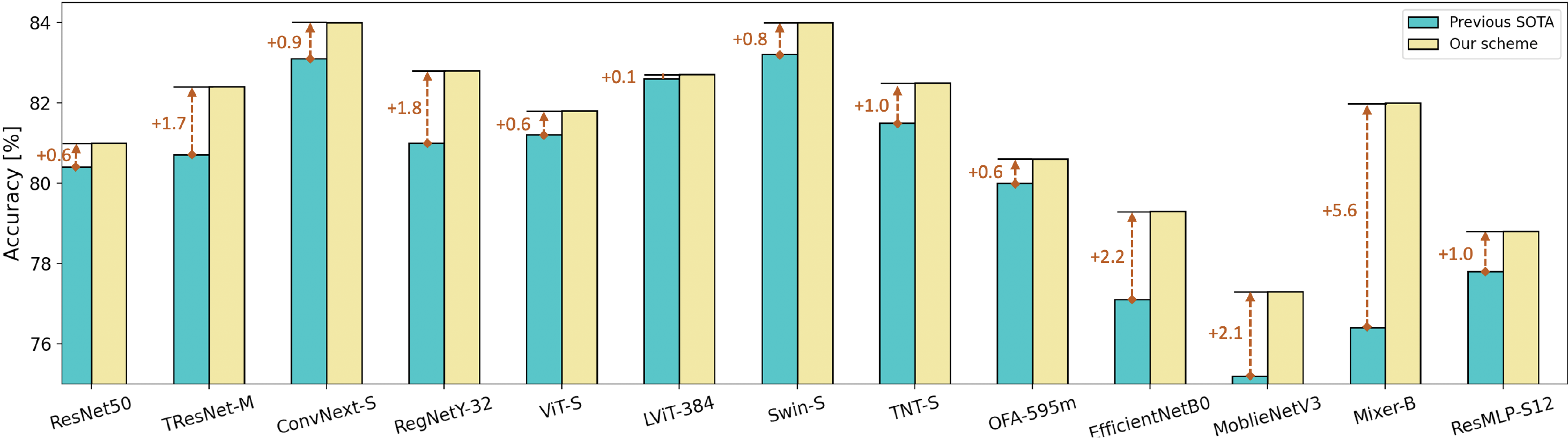}
  \caption{Top-1 accuracy comparison (\%)}
\end{subfigure}
% \caption{\textbf{A unified scheme proposal for various architectural families.} (a) We propose a unified scheme for training ImageNet with various architecture families. (b) Our scheme consists of a set of techniques (such as KD, EMA and strong augmentations) that were chosen based on their ability to generalize well to various architectures. (c) ImageNet accuracy (\%) performances of the pr  oposed scheme and a comparison to other reported baseline approaches for various architectures.}
\caption{\textbf{Our unified training scheme for ImageNet, \method{}}. With \method{}, we can train any backbone to top results on ImageNet, without any hyper-parameter tuning or adjustments per architecture.}
\label{fig:intro}
\end{figure*}

ImageNet (1K) dataset, introduced for the ILSVRC2012 visual recognition challenge \cite{ILSVRC15}, has been at the center of
modern advances in deep learning \cite{krizhevsky2012imagenet, he2016deep,sandler2018mobilenetv2}. It serves as a main dataset for pretraining computer-vision models \cite{tan2019efficientnet,lee2020compounding,howard2019searching}, and measuring the accuracy of models on ImageNet was found to be a good proxy to actual performance on  
various downstream tasks \cite{kornblith2019better,ridnik2021imagenet,wightman2021resnet}.

% presenting the basic problem 
However, training on ImageNet remains an ongoing challenge. Since the seminal work of AlexNet~\cite{krizhevsky2012imagenet}, new training tricks, regularizations and enhancements to improve results have been continuously suggested: One-cycle learning rate scheduling~\cite{smith2019super}; Stronger augmentations based on AutoAugment~\cite{cubuk2019autoaugment} and RandAugment~\cite{cubuk2020randaugment}; Scaling learning rate with batch size~\cite{you2017scaling}; Exponential-moving average (EMA) of model weights~\cite{izmailov2018averaging}; Improved weights initializations~\cite{hanin2018start,he2019bag}; Image-based regularizations such as Cutout~\cite{devries2017improved}, Cutmix~\cite{yun2019cutmix} and Mixup~\cite{zhang2017mixup}; Architecture regularizations like drop-path~\cite{cai2019effective} and drop-block~\cite{ghiasi2018dropblock}; Label-smoothing~\cite{muller2019does}; Different train-test resolutions~\cite{touvron2019fixing}; More training epochs~\cite{wightman2021resnet}; Progressive image resizing during training~\cite{tan2021efficientnetv2}; True weight decay~\cite{loshchilov2017decoupled}; Dedicated optimizer for large batch size~\cite{you2019large}, and more.

Almost any new architecture proposed is accompanied by a dedicated training scheme  (e.g. \cite{liu2022convnet,touvron2021training,ridnik2021imagenet,tan2021efficientnetv2,he2019bag}). 
These schemes can significantly differ, and a tailored scheme for one model often underperforms when used for other models. 
For example, when a dedicated scheme for ResNet50 model  \cite{wightman2021resnet} was used for training EfficieneNetV2 model, it yielded  $3.3\%$ lower accuracy than a tailor-made scheme \cite{tan2021efficientnetv2}.

Broadly speaking, deep learning backbones for computer vision can be divided into four categories: ResNet-like, Mobile-oriented, Transformers, and MLP-only.
ResNet model~\cite{he2016deep} and its variants (TResNet~\cite{ridnik2021tresnet}, SEResNet~\cite{hu2018squeeze}, ResNet-D~\cite{he2019bag} to name a few) usually work well over a wide variety of training schemes~\cite{ridnik2021imagenet}. A top-performance training scheme for ResNet models was suggested in \cite{wightman2021resnet}, and became the standard for this type of models.
% A typical ResNet-like training scheme usually uses SGD or Adam optimizer and one-cycle learning rate scheduling. 
%
Mobile-oriented models are architectures that heavily rely on depthwise convolutions, and efficient CPU-oriented design~\cite{sandler2018mobilenetv2,howard2019searching,tan2019efficientnet}.
Their dedicated training schemes usually consist of RMSProp optimizer, waterfall learning rate scheduling and EMA. 
Due to lack of inductive bias, Transformer-based~\cite{dosovitskiy2020image,liu2021swin,graham2021levit} and MLP-only~\cite{tolstikhin2021mlp,touvron2021resmlp} models for vision are harder to train, and less stable \cite{ridnik2021imagenet,liu2020understanding}. A dedicated training scheme for these models was proposed in \cite{touvron2021training}, which includes longer training (1000 epochs), strong cutmix-mixup and drop-path regularizations, large weight-decay and repeated augmentations~\cite{berman2019multigrain}.
%

% KD paragraph
Knowledge Distillation (KD), originally proposed in \cite{hinton2015distilling}, uses a teacher model to guide the target network (often referred to as the student model) along the training. The student receives supervision from both the ground-truth label and the teacher's prediction for each image. KL-divergence measures the extra loss between the teacher the and student.
\cite{touvron2021training} proposed a dedicated KD scheme for training Transformers-based models on ImageNet, that relies on a special distillation token, hard-label thresholding and long training.  \cite{cai2020once} used KD from super-network to train sub-networks during neural architecture search process. \cite{lee2020compounding} showed marginal improvement for ResNet50 model when using KD.
\cite{wei2020circumventing} demonstrated that KD suppresses noise from data augmentations, and enables using stronger augmentations. \cite{yun2021relabeling} suggested  a variant of KD by cropping the spatial prediction layer of the teacher.

However, KD is not a common practice for ImageNet training. Most training schemes  proposed throughout the years have not utilized KD, and it is not an integral part of the popular repositories for ImageNet training, such as \cite{rw2019timm}. For obtaining top results, a more frequent option is pretraining a model on the larger ImageNet-21K dataset~\cite{ridnik2021imagenet}, and fine-tuning it on ImageNet-1K ~\cite{liu2021swin,tan2021efficientnetv2,liu2022convnet,touvron2021resmlp,ridnik2021tresnet}. This alternative, of course, requires longer training and larger computational budget. It is a common practice to separate between results obtained from using only ImageNet-1K images, to results obtained using \textit{extra-data}, such as ImageNet-21K pretraining.

% \TBD{tbd: try to add more motivation why KD is so good. talk about it adds more information that the hard labels of the ground-truth. KD present correlations, similarities between classes, it fixes bad taggings, handles better pictures with several objects and so on. with better supervision, the entire optimization process is more effective and robust. This is especially important for imagenet, where many classes are similar (type of fishes). maybe add a picture to demonstrate what is good about KD predictions vs ground truth\\}

In this paper, we introduce a unified training scheme for ImageNet, called \method{} (Unified Scheme for ImageNet). \method{} can train any backbone to state-of-the-art results, \textbf{without any hyper-parameter tuning} or tailor-made tricks per model. An illustration of \method{} training scheme appears in Figure \ref{fig:intro}. The scheme is based on the observation that vanilla KD, which imposes an additional KL-divergence loss between the teacher and student predictions, works very well on any backbone.
% . (2) When using KD, the training process is far more robust, and requires fewer training tricks and regularizations.
%
The reason for the effectiveness of KD is the introduction of additional information, that does not exist in the original ground-truth labels. The teacher's predictions per image account for correlations and similarities between classes, they handle better pictures with several objects, and even compensate for ground-truth mistakes. KD also handle better augmentations, and removes the need for label smoothing. All these factors lead to a more robust and effective optimization process, that requires fewer training tricks and regularizations.

We thoroughly test \method{} on a wide variety of modern deep learning models, including ResNet-like, Mobile-oriented, Transformer-based and MLP models. On all model tested, \method{} outperforms previously reported state-of-the-art results, that were obtained with a dedicated scheme per model. \method{} is also efficient, and requires only $300$ epochs of training. 

Since the proposed scheme consistently leads to top results, it also enables a fair comparison of speed-accuracy trade-offs. We benchmark various models on GPU and CPU, and identify leading models that provide the best speed-accuracy trade-off along the Pareto curve.

The paper’s contributions can be summarized as follow:
\begin{itemize}[leftmargin=0.4cm]
  \setlength{\itemsep}{0.2pt}
  \setlength{\parskip}{0.2pt}
  \setlength{\parsep}{0.2pt}
  \item We introduce a unified, efficient training scheme for ImageNet dataset, \method{}, that does not require hyper-parameter tuning. Exactly the same recipe is applied to any backbone. Hence, ImageNet training is transformed from an expert-oriented task into an automatic seamless procedure.
  \item We test \method{} on various deep learning models, including ResNet-like, Mobile-oriented, Transformer-based and MLP-only models. We show it consistently and reliably achieves state-of-the-art results, compared to tailor-made schemes per model.
  \item We use \method{} to perform a methodological speed-accuracy comparison of modern deep learning models, and identify efficient backbones along the Pareto curve.
\end{itemize}

\begin{figure*}[t!]
\centering
  \includegraphics[scale=0.58]{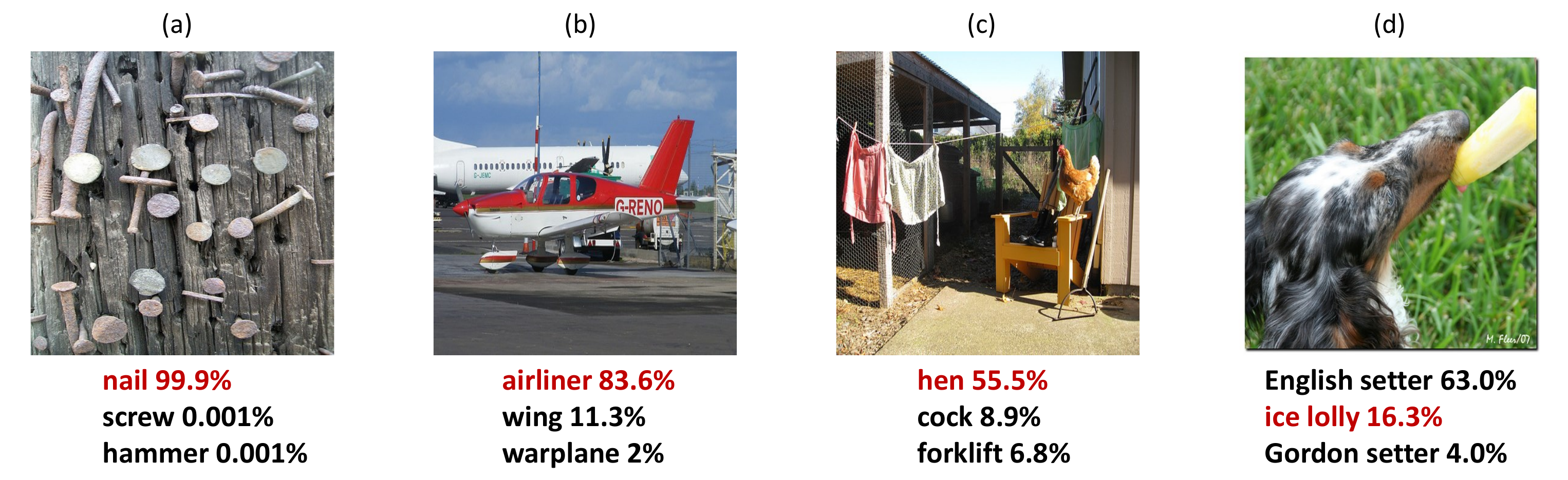}
  \caption{\textbf{Examples for teacher predictions}. ImageNet ground-truth labels are highlighted in red. Unlike the ground-truth, the teacher predictions account for similarities and correlations between classes, objects' saliency, pictures with several objects, and more. The teacher predictions would also better represent the content of images under strong augmentations.}
\label{fig:kid_vs_gt}
\vspace{-0.2cm}
\end{figure*}

\section{Method}
In this section, we will first review how to apply knowledge distillation (KD) for classification. Then we will discuss why we need KD in ImageNet training, and finally present USI, our KD-based training scheme for ImageNet.

\subsection{KD for Classification}
For any input image, a classification network outputs a logit vector $\mathbf{z} = \{z_i\}_{i=1}^K$, where $K$ is the number of classes.
The softened prediction vector is denoted by $\mathbf{p}(\tau) = \{p_i\}_{i=1}^K$, where each element $p_i(\tau)$ is given by applying the softmax activation function, with a temperature-scaling parameter $\tau$:
\begin{equation}
p_i(\tau)= \frac {\exp({z_i/\tau})}{\sum_{j=1}^{K} \exp({z_j/\tau})}.
\end{equation}
Let us define the softened prediction vectors of the  student and teacher models as $\mathbf{p}^{s}(\tau)$ and $\mathbf{p}^{t}(\tau)$, respectively. 
The objective function used for training the student model is a combination of the Cross-Entropy (CE) loss between the student's prediction and the ground-truth vector $\mathbf{y}$, and the Kullback-Leibler (KL) divergence between the student and the teacher predictions,
\begin{equation}\label{eq:KD_loss}
    \mathcal{L} = \mathcal{L}_{\text{CE}}(\mathbf{p}^{s}(1), \mathbf{y}) + \alpha_{\text{kd}} \mathcal{L}_{\text{KL}}(\mathbf{p}^{s}(\tau), \mathbf{p}^{t}(\tau)),
\end{equation}
where $\alpha_{kd}$ is a hyper-parameter for adjusting the relative importance of the KD loss. 
The CE loss is given by,
\begin{equation}
    \mathcal{L}_{\text{CE}}(\mathbf{p}^{s}(1), \mathbf{y}) = -\sum_{j} y_j \log p^{s}_j(1),
\end{equation}
and the KL divergence loss is given by,
\begin{equation}
    \mathcal{L}_{\text{KL}}(\mathbf{p}^{s}(\tau), \mathbf{p}^{t}(\tau)) = \tau^2 \sum_{j} p^{t}_j(\tau) \log \frac{p^{t}_j(\tau)}{p^{s}_j(\tau)}.
\end{equation}

\subsection{Why Do We Need KD in ImageNet Training ?}
\label{method:why_kd}
 ImageNet serves as the primary dataset for pretraining and evaluating computer-vision models. Unlike other classification datasets, on ImageNet we are training models from \textit{scratch}, and not doing \textit{transfer learning}. Training from scratch is, in general, harder, and requires higher learning rates, stronger regularizations, and more training epochs. Hence, the optimization process on ImageNet is more sensitive to different hyper-parameters, and the architecture used.

To gain more insight and motivation into the impact of KD, we present in Figure \ref{fig:kid_vs_gt} some typical examples for the predictions of the teacher model, compared to ImageNet ground-truth labels.
\begin{itemize}[leftmargin=0.4cm]
  \setlength{\itemsep}{0.2pt}
  \setlength{\parskip}{0.2pt}
  \setlength{\parsep}{0.2pt}
  \item Picture (a) contains big salient \textit{nails}. That is the ground-truth, and that's also the teacher leading prediction ($99.9\%$). Notice that the teacher's 2nd and 3rd top predictions are related to nails (\textit{screw} and \textit{hammer}), but with negligible probability.
  \item Picture (b) contains an \textit{airliner}. That is the top prediction of the teacher ($83.6\%$). However, the teacher also predicts \textit{wing} with a non-negligible probability ($11.3\%$). That is not a mistake - an airliner has wings. The teacher here mitigates the case where the ground-truth labels are not mutually-exclusive, and provides more accurate information about the content of the image.
  \item Picture (c) contains a \textit{hen} (female chicken). However, the hen is not very big and salient. The teacher's predictions reflects that, by identifying an hen with lower probability ($55.5\%$). The teacher also gives non-negligible probability to \textit{cock} (male chicken) label - $8.9\%$. This is a mistake by the teacher, but a logical one - hen and cock are quite similar.
   \item In picture (d) the teacher disagrees with the ground-truth. The ground-truth is \textit{ice-lolly}, while the teacher top prediction is \textit{English setter} (a type of dog). The teacher is correct - the dog is more salient in the picture.
\end{itemize}
We see from the above examples that the teacher's predictions contain more information than the plain (single-label) ground-truth. The  rich predictions provided by the teacher account for correlations and similarities between classes. They handle better pictures with several objects, and even compensate for ground-truth mistakes.  
KD predictions also handles better strong augmentations, since  they represent the correct content of the augmented image. They also removes the need for label smoothing, since the teacher inherently outputs soft predictions.

Due to these factors, training with a teacher provides better supervision, leading to a more effective and robust optimization process, compared to training with hard-labels only. 
% The improved optimization process enables us to use a unified scheme on ImageNet, that works on any backbone.

\subsection{The Proposed Training Scheme}
\label{method:training_scheme}
Our proposed training scheme for ImageNet, \method{}, is based on utilizing KD. When training on ImageNet with KD, we observe that the training process is far more robust to hyper-parameter selection, and requires fewer training tricks and regularizations. In addition, the need for dedicated tricks per backbone is eliminated - a single unified scheme can train any backbone to top results. 
An illustration of \method{} scheme appears in Figure \ref{fig:intro}. In Table \ref{Table:train_config} we present the full training configuration.
\begin{table}[hbt!]
\centering
\begin{tabular}{c|c} 
\Xhline{3\arrayrulewidth}
Procedure        & Value               \\ 
\Xhline{3\arrayrulewidth}
Train resolution & 224                 \\ 
Test resolution & 224                 \\ 
Epochs           & 300                 \\
Optimizer        & AdamW               \\
Weight decay     & 2e-2                \\
Learning rate    & 2e-3                \\
LR decay         & One-cycle policy    \\
Mixup alpha      & 0.8                 \\
Cutmix alpha     & 1.0                 \\
Augmentations    & Rand-augment (7/0.5)  \\
Test crop ratio  & 0.95                \\ 
Repeated Augs  & 3                \\ 
\hline
Base loss        & Cross entropy       \\
KD loss          & KL-divergence       \\
KD temperature   & 1                   \\
$\alpha_{kd}$    & 5                   \\ 
Teacher         & TResNet-L                   \\ 

\hline
Batch size       & 512 to 3456           \\
\hline
\end{tabular}
\caption{\textbf{\method{} training configuration}. With \method{}, exactly the same training recipe is applied to any backbone, and no hyper-parameter tuning is needed.}
\label{Table:train_config}
\vspace{-0.2cm}
\end{table}
\\
Some observations and insights into the proposed scheme:
\paragraph{Batch size selection}
The maximal batch size allowed by Different backbones varies significantly (see Table \ref{Table:model_details} in the appendix). Hence, using a fixed batch size for all backbones is not always possible.
It is beneficial to choose a batch size as large as possible, since it enables fully utilizing the GPU cores, reducing communication overheads, and increasing training speed.
Previous schemes suggested that larger batches require larger learning rates or a dedicated optimizer~\cite{you2017scaling,gotmare2018closer}.

\method{}, which is KD-based training scheme with AdamW optimizer, is more robust to batch size and learning rate tuning. We will show in Section \ref{results:batch_size} that with the same learning rate, \method{} consistently provides top results for a wide range of batch sizes. Hence, in Table \ref{Table:train_config} we state a range of batch sizes, instead of a fixed one. Any value along this range can be chosen. We recommend using $0.8$ to $0.9$ of the maximal possible batch size possible, for optimizing training speeds.

\paragraph{Which teacher to choose} Our primary requirement is to choose a teacher who outperforms the student, a common requirement in KD~\cite{hinton2015distilling}. Given that constraint, we suggest selecting a teacher model with good speed-accuracy trade-off (see Figure \ref{fig:speed_acc_v100}). We will show in Section \ref{results:teacher_type} that our scheme is robust to teacher and student types. Teachers with similar accuracy train students to similar accuracy, regardless of their type (CNNs or Transformers).

% we put the table here just do to graphical issues
\begin{table*}[hbt!]
\centering
\begin{tabular}{|c|c|c|cccc|} 
\Xhline{3\arrayrulewidth}
\multirow{2}{*}{\begin{tabular}[c]{@{}c@{}}Model \\Type\end{tabular}}            & \multirow{2}{*}{\begin{tabular}[c]{@{}c@{}}Model \\Name\end{tabular}} & \multirow{2}{*}{\begin{tabular}[c]{@{}c@{}}\method{} \\Top1 Acc. [\%]\end{tabular}} & \multicolumn{4}{c|}{Comparable Training Scheme}                     \\ 
\cline{4-7}
                                                                                 &                                                                       &                                                                                      & Top1 Acc. [\%] & Epochs    & KD  & Additional Details             \\ 
\Xhline{3\arrayrulewidth}
\multirow{5}{*}{CNN}                                                             & \multirow{2}{*}{ResNet50}                                             & \multirow{2}{*}{81.0}                                                                & 80.4~\cite{wightman2021resnet}           & 600       & no  & ResNet-strike-back, A1 config  \\
                                                                                 &                                                                       &                                                                                      & 80.2~\cite{yun2021relabeling}           & 300       & yes & Relabel-based KD                   \\ 
\cline{2-7}
                                                                                 & TResNet-M                                                              & 82.4                                                                          & 80.7 \cite{ridnik2021tresnet}          & 300       & no  &                                \\ 
\cline{2-7}
                                                                                 & ConvNext-S                                                            & 84.0                                                                         & 83.1 \cite{liu2022convnet}           & 300       & no  &                                \\ 
\cline{2-7}
                                                                                 & RegNetY-32                                                            &    82.8                                                                                  & 81.0 \cite{radosavovic2020designing}           & 100       & no  &                                \\ 
\Xhline{3\arrayrulewidth}
\multirow{5}{*}{Transformer}                                                     & \multirow{2}{*}{ViT-S}                                                & \multirow{2}{*}{81.8}                                                                & 79.8~\cite{dosovitskiy2020image}           & 300       & no  & Original paper scheme          \\
                                                                                 &                                                                       &                                                                                      & 81.2~\cite{touvron2021training}           & 1000      & yes & DeiT scheme                    \\ 
\cline{2-7}
                                                                                 & LeViT-384                                                             & 82.7                                                                        & 82.6~\cite{graham2021levit}            & 1000      & yes & DeiT scheme   \\ 
\cline{2-7}
                                                                                 & Swin-S                                                                & 84.0                                                                       & 83.2~\cite{liu2021swin}           & 300       & no  &                                \\ 
\cline{2-7}
                                                                                 & TNT-S                                                                 & 82.5                                                                      & 81.5~\cite{han2021transformer}           & 300       & no  &                                \\                                                                                  
\Xhline{3\arrayrulewidth}
\multirow{3}{*}{\begin{tabular}[c]{@{}c@{}}Mobile-~\\Oriented\\CNN\end{tabular}} & OFA-595m                                                              & 80.6                                                                         & 80.0~\cite{cai2020once}           & 255       & yes & KD from super network          \\ 
\cline{2-7}
                                                                                 & EfficientnetB0                                                     &      79.3                                                                               & 77.1~\cite{tan2019efficientnet}        & not stated & no  &                                \\ 
\cline{2-7}
                                                                                 & MobileNetV3                                                          &                                                                           77.3          & 75.2~\cite{howard2019searching}                & not stated & no  & \multicolumn{1}{l|}{}          \\

\Xhline{3\arrayrulewidth}
\multirow{2}{*}{MLP-Only} & 
Mixer-B                                                          & 82.0                                                                         & 76.4~\cite{tolstikhin2021mlp}           & 255       & no &        \\ 

\cline{2-7} & ResMLP-S12                                                          & 78.8                                                                         & 77.8~\cite{touvron2021resmlp}           & 400       & yes &        \\ 
\Xhline{3\arrayrulewidth}

\end{tabular}
\caption{Comparison of our proposed scheme, \method{}, to previous state-of-the-art results. Train and test resolution - $224$. }
\label{Table:main_result}
\vspace{-0.2cm}
\end{table*}

\paragraph{KD impact on training speed}
Adding KD supervision incurs additional overhead, and reduces training speed. However, the additional overhead is usually small. while the student network needs to do forward pass, store intermediate maps, do backward pass and update weights, the teacher network only needs forward passes. In addition, since the teacher model is fixed, we can apply various optimizations to it, like batch-norm fusion, channels-last and jit~\cite{ridnik2021tresnet}.
We found that the relative overhead of KD decreases as we increase the batch size, which is another reason to prefer large batch sizes.
For TResNet-L~\cite{ridnik2021tresnet} teacher model ($83.9\%$ accuracy), the additional overhead from KD reduces the training speed by $10\%$-$20\%$.

\section{Results}
\label{sec:results}
\subsection{Comparison to Previous Schemes}
In Table \ref{Table:main_result} we present the ImageNet top-1 accuracy obtained for various deep learning architectures, when trained with our proposed scheme, \method{}. We compare USI to previous state-of-the-art results, obtained with a tailor-made scheme per architecture.
We see from Table \ref{Table:main_result} that on all architecture tested (CNN, Transformer, Mobile-oriented, MLP-only), \method{} reaches results better than previously reported top results:

\begin{itemize}[leftmargin=0.4cm]
  \setlength{\itemsep}{0.2pt}
  \setlength{\parskip}{0.2pt}
  \setlength{\parsep}{0.2pt}
  \item For CNN architectures, \method{} significantly outperforms previous results. This applies also to ResNet50 model, where we compare USI to a recently proposed dedicated scheme~\cite{wightman2021resnet}, and a KD-based scheme~\cite{yun2021relabeling}.
  \item For Transformer architectures, on two prominent models, ViT-S and LeViT-384, \method{} reaches better results compared to DeiT scheme~\cite{touvron2021training}, that also utilized KD. Notice that DeiT used a tailored KD training recipe for Transformers, which includes longer training ($1000$ epochs), an auxiliary KD head from special distillation token, and modifications to the KD algorithm (hard-thresholding the teacher predictions). In contrast, \method{} uses vanilla KD, less than a third of the epochs, and no architecture-tailored modifications. With equal number of epochs, the gap between \method{} and DeiT is even bigger (see Section \ref{results:epochs}).
  \item For Mobile-oriented and MLP-based architectures, USI shows significant improvement compared to previously reported results.
\end{itemize}
To conclude, Table \ref{Table:main_result} presents results for a wide variety of modern deep learning models. For each model, we compare results from our proposed scheme, \method{}, to top results from the literature, obtained with tailor-made schemes per model. On all models tested, without any hyper-parameter tuning, \method{} achieves state-of-the-art results. We believe that this is a fair comparison, that demonstrates that \method{} enables to transform high-quality training on ImageNet into an automatic routine, that requires no dedicated expertise and cumbersome tuning process.

\subsection{Robustness to Batch-size}
\label{results:batch_size}
As discussed in Section \ref{method:training_scheme}, larger batch size leads to faster training speed. 
% Previous works suggested that to obtain top results, as we increase the batch size the learning rate should also be scaled.
%
In Table \ref{Table:batch_size} we test the robustness of \method{}, with fixed learning rate, to different batch sizes. For the tests we used TResNet-M model, which enables large maximal batch size due to usage of inplace-activated batch-norm instead of regular batch-norm~\cite{ridnik2021tresnet}
Table \ref{Table:batch_size} demonstrates that over a large range of batch sizes, $512$-$3456$, the accuracy remains almost the same.  This indicates that  \method{} operates well with a fixed learning rate. 
\begin{table}[hbt!]
\centering
\begin{tabular}{c|c|c} 
\Xhline{3\arrayrulewidth}
\begin{tabular}[c]{@{}c@{}}Batch \\Size\end{tabular} & \begin{tabular}[c]{@{}c@{}}Top1 Acc. \\{[}\%]\end{tabular} & \begin{tabular}[c]{@{}c@{}}Training speed \\{[}img/sec]\end{tabular}  \\ 
\Xhline{3\arrayrulewidth}
512                                                  & 82.3                                                       & 1100                                                                  \\
1024                                                 & 82.5                                                       & 1900                                                                  \\
2048                                                 & 82.3                                                       & 3000                                                                  \\
2752                                                 & 82.4                                                       & 4100                                                                  \\
3456                                                 & 82.4                                                       & 4300                                                                  \\
\hline
3456                                                 &                                                     & {\begin{tabular}[c]{@{}c@{}}4900 \\(no-KD reference)\end{tabular}}                                                                \\
\hline
\end{tabular}
\caption{Accuracy and training speed, for different batch sizes. Model tested - TResNet-M.}
\label{Table:batch_size}
\vspace{-0.2cm}
\end{table}

%
% Table \ref{Table:batch_size} demonstrates that over a large range of batch sizes, $512$-$3456$, the accuracy remains almost the same. 
% This demonstrates that our proposed scheme can operate with a fixed learning rate for a wide range of batch sizes. 
% This indicates that  \method{} operates well with a fixed learning rate. 
%

For our runs, we used an 8xV100 Nvidia machine, TResNet-M student, and TResNet-L teacher. In terms of training speed, with a batch size of $512$ ($64$ per GPU) we reached a training speed of $1100$ img/sec, while with a batch size of $3456$ ($432$ per GPU) we reached $4300$ img/sec. Hence, increasing the batch size significantly aids the training speed. Without KD, we reached a training speed of $4900$ img/sec.

\subsection{Robustness to Teacher Type}
\label{results:teacher_type}
The results in Table \ref{Table:main_result} were obtained with TResNet-L teacher. We chose this model as a teacher since it provides high accuracy, and a good speed-accuracy trade-off on GPU (see Figure \ref{fig:speed_acc_v100}).
In Table \ref{Table:teacher_type} we test if different student models benefit from different teacher models. Since TResNet-L is a CNN, we compared it to a Transformer-based teacher, Volo-d1~\cite{yuan2021volo}, which has similar top-1 accuracy ($83.9\%$ for TResNet-L, $84.1\%$ for Volo-d1).

\begin{table}[hbt!]
% \scalebox{0.94}{
\resizebox{\columnwidth}{!}{%
\begin{tabular}{c|c|c|c|c} 
\Xhline{3\arrayrulewidth}
Student                   & \begin{tabular}[c]{@{}c@{}}Student \\Type\end{tabular} & Teacher   & \begin{tabular}[c]{@{}c@{}}Teacher \\Type\end{tabular} & \begin{tabular}[c]{@{}c@{}}Top1 \\Acc. [\%]\end{tabular}  \\ 
\Xhline{3\arrayrulewidth}
\multirow{2}{*}{ResNet50} & \multirow{2}{*}{CNN}                                   & TResNet-L & CNN                                                    & 81.0                                                  \\ 
\cline{3-5}
                          &                                                        & Volo-d1   & Transformer                                            & 80.9                                                  \\ 
\hline
\multirow{2}{*}{LeViT384} & \multirow{2}{*}{Transformer}                           & TResNet-L & CNN                                                    & 82.7                                                  \\ 
\cline{3-5}
                          &                                                        & Volo-d1   & Transformer                                            & 82.7                                                  \\
\hline
\end{tabular}}
\caption{Testing different students with different teachers.}
\label{Table:teacher_type}
\end{table}

We see from Table \ref{Table:teacher_type}  that both CNN and Transformer students work well with CNN and Transformer teachers. This implies we have flexibility in choosing the teacher type. 
% We recommend selecting a teacher with fast inference speed.

\subsection{Number of Training Epochs}
\label{results:epochs}
\method{} default configuration is KD training, for $300$ epochs. However, for reaching the maximal possible accuracy, $300$ epochs is not enough, and longer training would further improve the accuracy. This phenomenon, where in KD the student model benefits from very long training, is called \textit{patient teacher}~\cite{beyer2021knowledge}.
In Table \ref{Table:trainin_epochs} we present the accuracies obtained for various training lengths.
\begin{table}[hbt!]
\centering

\begin{tabular}{c|c} 
\Xhline{3\arrayrulewidth}
Training epochs & Top1 Acc. [\%]  \\ 
\Xhline{3\arrayrulewidth}

100         & 80.0             \\
200         & 81.9             \\
300      & 82.7             \\
600       & 83.0   \\
1000       & 83.2 \\
\hline
\end{tabular}
\caption{Accuracy for  different numbers of epochs. Model tested - LeViT-384.}
\label{Table:trainin_epochs}
\vspace{-0.2cm}
\end{table}
As can be seen, the accuracy continues to improve as we increase training epoch from $300$ to $600$ and to $1000$.

Our default training configuration was chosen to be $300$ epochs since this value provides a good compromise - training times are reasonable ($1$-$3$ days on 8xV100 GPU, depending on the model), while we consistently achieve good results, as can be seen in Figure~\ref{fig:intro} (c). However, if training times are not a limitation, we recommend increasing the number of training epochs.

\subsection{Speed-Accuracy Measurements}
When using a tailor-made training scheme per backbone, that depends on hyper-parameter tuning, computing power and other factors, it is challenging to reliably compare different models.
With \method{}, that consistently provides top results for any backbone, we can do a methodological, reproducible and reliable comparison of speed-accuracy trade-off on ImageNet.
In Figure \ref{fig:speed_acc_tensorrt_v100} and Figure \ref{fig:speed_acc_cpu} we compare various modern architectures, on GPU and CPU inference.

% should be at the top of page 7
% \input{speed_acc_figure_v100}
\pgfplotsset{grid style={dashed}}
\begin{figure*}
    \centering
    \begin{tikzpicture}
    \definecolor{darkgreen}{rgb}{0.0, 0.5, 0.0}
    % \pgfplotscreateplotcyclelist{mycolorlist}{%
    % blue,every mark/.append style={fill=blue!80!black},mark=*\\%
    % red,every mark/.append style={fill=red!80!black},mark=square*\\%
    % brown!60!black,every mark/.append style={fill=brown!80!black},mark=otimes*\\%
    % black,mark=star\\%
    % blue,every mark/.append style={fill=blue!80!black},mark=diamond*\\%
    % red,densely dashed,every mark/.append style={solid,fill=red!80!black},mark=*\\%
    % brown!60!black,densely dashed,every mark/.append style={
    % solid,fill=brown!80!black},mark=square*\\%
    % black,densely dashed,every mark/.append style={solid,fill=gray},mark=otimes*\\%
    % blue,densely dashed,mark=star,every mark/.append style=solid\\%
    % red,densely dashed,every mark/.append style={solid,fill=red!80!black},mark=diamond*\\%
    % }
    \begin{axis}[
    	xlabel={Inference Speed [images/sec]},
    	ylabel={Top-1 Accuracy [\%]},
    % 	cycle list name=mycolorlist,
    	width=\textwidth*0.9,
    	height=\axisdefaultwidth*0.87,
        legend style={anchor=north east},
    	legend entries={Transformer,
                CNN,
                Mobile,
                MLP-Only},
        xmin=1000,
        xmax=11000,
        grid=major,
        scaled x ticks=false,
    ]
    \addlegendentry[mark=square*, color=blue]{Transformer}
    \addlegendentry[mark=*, color=red]{CNN}
    \addlegendentry[mark=triangle*, color=orange]{Mobile-Oriented}
    \addlegendentry[mark=star*, color=darkgreen]{MLP-only}
    
    \addplot [mark=square*, color=blue,
        visualization depends on=\thisrow{alignment} \as \alignment,
        nodes near coords, % Place nodes near each coordinate
        point meta=explicit symbolic, % The meta data used in the nodes is not explicitly provided and not numeric
        every node near coord/.style={anchor=\alignment} % Align each coordinate at the anchor 40 degrees clockwise from the right edge
        ] table [% Provide data as a table
         meta index=2 % the meta data is found in the third column
         ] {
    x       y       label       alignment
    2800    84.2    \textbf{LeViT-768*}     180
    };
    
    \addplot [mark=square*,  color=blue,
        visualization depends on=\thisrow{alignment} \as \alignment,
        nodes near coords, % Place nodes near each coordinate
        point meta=explicit symbolic, % The meta data used in the nodes is not explicitly provided and not numeric
        every node near coord/.style={anchor=\alignment} % Align each coordinate at the anchor 40 degrees clockwise from the right edge
        ] table [% Provide data as a table
         meta index=2 % the meta data is found in the third column
         ] {
    x       y       label       alignment
    6461.44    82.7    \textbf{LeViT-384}      220
    };

    \addplot [mark=asterisk, color=red,
        visualization depends on=\thisrow{alignment} \as \alignment,
        nodes near coords, % Place nodes near each coordinate
        point meta=explicit symbolic, % The meta data used in the nodes is not explicitly provided and not numeric
        every node near coord/.style={anchor=\alignment} % Align each coordinate at the anchor 40 degrees clockwise from the right edge
        ] table [% Provide data as a table
         meta index=2 % the meta data is found in the third column
         ] {
    x       y       label       alignment
    1336.32    84    ConvNext-S      200
    };
    \addplot [mark=triangle*, color=orange,
        visualization depends on=\thisrow{alignment} \as \alignment,
        nodes near coords, % Place nodes near each coordinate
        point meta=explicit symbolic, % The meta data used in the nodes is not explicitly provided and not numeric
        every node near coord/.style={anchor=\alignment} % Align each coordinate at the anchor 40 degrees clockwise from the right edge
        ] table [% Provide data as a table
         meta index=2 % the meta data is found in the third column
         ] {
    x       y       label       alignment
    7577.6    80.6    OFA-595m      180
    };
    \addplot [mark=o, color=darkgreen,
        visualization depends on=\thisrow{alignment} \as \alignment,
        nodes near coords, % Place nodes near each coordinate
        point meta=explicit symbolic, % The meta data used in the nodes is not explicitly provided and not numeric
        every node near coord/.style={anchor=\alignment} % Align each coordinate at the anchor 40 degrees clockwise from the right edge
        ] table [% Provide data as a table
         meta index=2 % the meta data is found in the third column
         ] {
    x       y       label       alignment
    6671.36    78.8    ResMLP-S12      180
    };
    \addplot [mark=square*, color=blue,
        visualization depends on=\thisrow{alignment} \as \alignment,
        nodes near coords, % Place nodes near each coordinate
        point meta=explicit symbolic, % The meta data used in the nodes is not explicitly provided and not numeric
        every node near coord/.style={anchor=\alignment} % Align each coordinate at the anchor 40 degrees clockwise from the right edge
        ] table [% Provide data as a table
         meta index=2 % the meta data is found in the third column
         ] {
    x       y       label       alignment
    1574.4    84    Swin-S      -270
    };
    \addplot [mark=asterisk, color=red,
        visualization depends on=\thisrow{alignment} \as \alignment,
        nodes near coords, % Place nodes near each coordinate
        point meta=explicit symbolic, % The meta data used in the nodes is not explicitly provided and not numeric
        every node near coord/.style={anchor=\alignment} % Align each coordinate at the anchor 40 degrees clockwise from the right edge
        ] table [% Provide data as a table
         meta index=2 % the meta data is found in the third column
         ] {
    x       y       label       alignment
    5608.96    82.4    TResNet-M      180
    };
    \addplot [mark=asterisk, color=red,
        visualization depends on=\thisrow{alignment} \as \alignment,
        nodes near coords, % Place nodes near each coordinate
        point meta=explicit symbolic, % The meta data used in the nodes is not explicitly provided and not numeric
        every node near coord/.style={anchor=\alignment} % Align each coordinate at the anchor 40 degrees clockwise from the right edge
        ] table [% Provide data as a table
         meta index=2 % the meta data is found in the third column
         ] {
    x       y       label       alignment
    3371.52    83.9    TResNet-L      180
    };
    \addplot [mark=square*, color=blue,
        visualization depends on=\thisrow{alignment} \as \alignment,
        nodes near coords, % Place nodes near each coordinate
        point meta=explicit symbolic, % The meta data used in the nodes is not explicitly provided and not numeric
        every node near coord/.style={anchor=\alignment} % Align each coordinate at the anchor 40 degrees clockwise from the right edge
        ] table [% Provide data as a table
         meta index=2 % the meta data is found in the third column
         ] {
    x       y       label       alignment
    3545.6    81.8    ViT-S      90
    };
    \addplot [mark=square*,  color=blue,
        visualization depends on=\thisrow{alignment} \as \alignment,
        nodes near coords, % Place nodes near each coordinate
        point meta=explicit symbolic, % The meta data used in the nodes is not explicitly provided and not numeric
        every node near coord/.style={anchor=\alignment} % Align each coordinate at the anchor 40 degrees clockwise from the right edge
        ] table [% Provide data as a table
         meta index=2 % the meta data is found in the third column
         ] {
    x       y       label       alignment
    10562.56    81.5    \textbf{LeViT-256}      0
    };
    \addplot [mark=asterisk, color=red,
        visualization depends on=\thisrow{alignment} \as \alignment,
        nodes near coords, % Place nodes near each coordinate
        point meta=explicit symbolic, % The meta data used in the nodes is not explicitly provided and not numeric
        every node near coord/.style={anchor=\alignment} % Align each coordinate at the anchor 40 degrees clockwise from the right edge
        ] table [% Provide data as a table
         meta index=2 % the meta data is found in the third column
         ] {
    x       y       label       alignment
    7808    80.9    ResNet-50      190
    };
    \addplot [mark=asterisk, color=red,
        visualization depends on=\thisrow{alignment} \as \alignment,
        nodes near coords, % Place nodes near each coordinate
        point meta=explicit symbolic, % The meta data used in the nodes is not explicitly provided and not numeric
        every node near coord/.style={anchor=\alignment} % Align each coordinate at the anchor 40 degrees clockwise from the right edge
        ] table [% Provide data as a table
         meta index=2 % the meta data is found in the third column
         ] {
    x       y       label       alignment
    4131.84    82.8    RegNetY-32      290
    };

    % \addplot [mark=triangle*, color=orange,
    %     visualization depends on=\thisrow{alignment} \as \alignment,
    %     nodes near coords, % Place nodes near each coordinate
    %     point meta=explicit symbolic, % The meta data used in the nodes is not explicitly provided and not numeric
    %     every node near coord/.style={anchor=\alignment} % Align each coordinate at the anchor 40 degrees clockwise from the right edge
    %     ] table [% Provide data as a table
    %      meta index=2 % the meta data is found in the third column
    %      ] {
    % x       y       label       alignment
    % 17044.48    77.3    MobileNetV3-L      -30
    % };
    
    \addplot [mark=triangle*, color=orange,
        visualization depends on=\thisrow{alignment} \as \alignment,
        nodes near coords, % Place nodes near each coordinate
        point meta=explicit symbolic, % The meta data used in the nodes is not explicitly provided and not numeric
        every node near coord/.style={anchor=\alignment} % Align each coordinate at the anchor 40 degrees clockwise from the right edge
        ] table [% Provide data as a table
         meta index=2 % the meta data is found in the third column
         ] {
    x       y       label       alignment
    7636.48    79.3    EfficientNet-B0      180
    };
    \addplot [mark=square*, color=blue,
        visualization depends on=\thisrow{alignment} \as \alignment,
        nodes near coords, % Place nodes near each coordinate
        point meta=explicit symbolic, % The meta data used in the nodes is not explicitly provided and not numeric
        every node near coord/.style={anchor=\alignment} % Align each coordinate at the anchor 40 degrees clockwise from the right edge
        ] table [% Provide data as a table
         meta index=2 % the meta data is found in the third column
         ] {
    x       y       label       alignment
    1400    82.5    TnT-S      180
    };
    \addplot [mark=o, color=darkgreen,
        visualization depends on=\thisrow{alignment} \as \alignment,
        nodes near coords, % Place nodes near each coordinate
        point meta=explicit symbolic, % The meta data used in the nodes is not explicitly provided and not numeric
        every node near coord/.style={anchor=\alignment} % Align each coordinate at the anchor 40 degrees clockwise from the right edge
        ] table [% Provide data as a table
         meta index=2 % the meta data is found in the third column
         ] {
    x       y       label       alignment
    1943.04    82    Mixer-B      -290
    };
    
    \addplot [mark=triangle*, color=orange,
        visualization depends on=\thisrow{alignment} \as \alignment,
        nodes near coords, % Place nodes near each coordinate
        point meta=explicit symbolic, % The meta data used in the nodes is not explicitly provided and not numeric
        every node near coord/.style={anchor=\alignment} % Align each coordinate at the anchor 40 degrees clockwise from the right edge
        ] table [% Provide data as a table
         meta index=2 % the meta data is found in the third column
         ] {
    x       y       label       alignment
    3979.7    82.1      EfficientNet-B3      178
    };

    % \legend{EfficientNet,TResNet}
    \end{axis}
    % \matrix [draw,below left] at (current bounding box.north east) {
    %   \node [blue,label=right:Transformer] {}; \\
    % };
   
    \end{tikzpicture}
    \caption{Speed-Accuracy comparison on an Nvidia V100 GPU, using TensorRT inference engine.}
    \label{fig:speed_acc_tensorrt_v100}
\end{figure*}
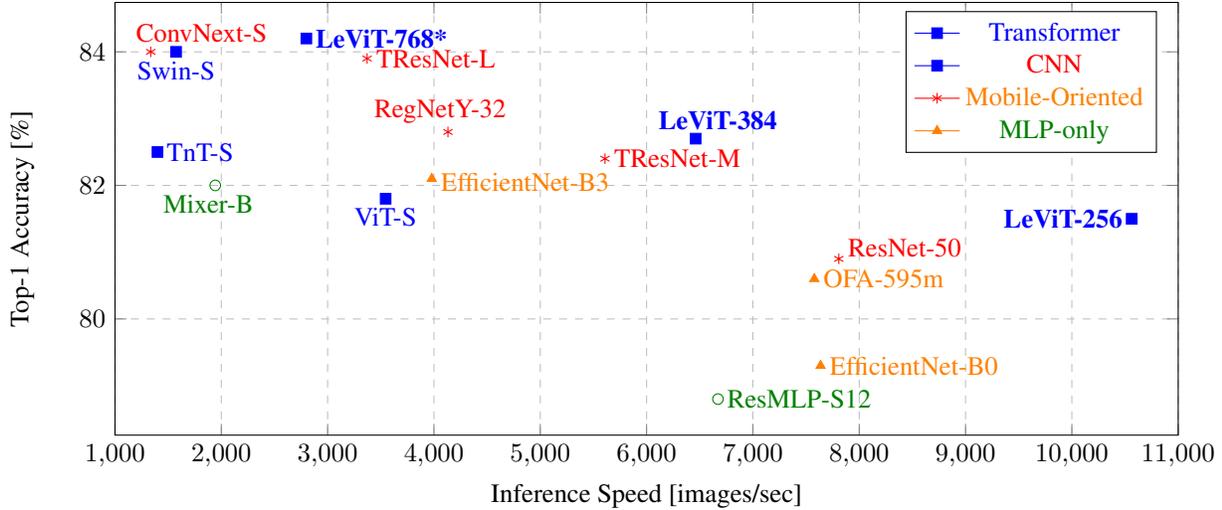
\pgfplotsset{grid style={dashed}}
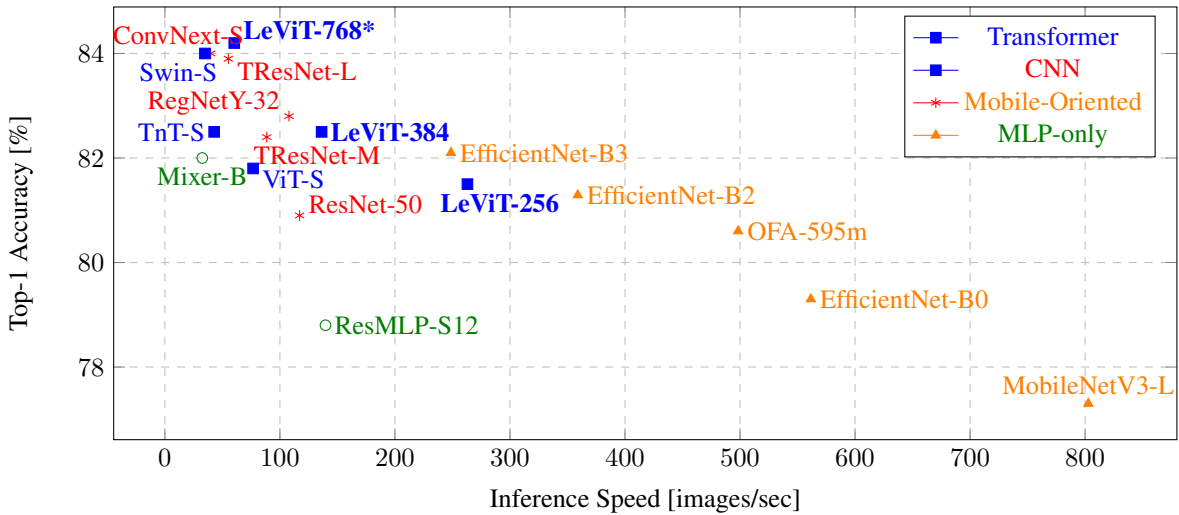
\begin{figure*}
    \centering
    \begin{tikzpicture}
    \definecolor{darkgreen}{rgb}{0.0, 0.5, 0.0}
    % \pgfplotscreateplotcyclelist{mycolorlist}{%
    % blue,every mark/.append style={fill=blue!80!black},mark=*\\%
    % red,every mark/.append style={fill=red!80!black},mark=square*\\%
    % brown!60!black,every mark/.append style={fill=brown!80!black},mark=otimes*\\%
    % black,mark=star\\%
    % blue,every mark/.append style={fill=blue!80!black},mark=diamond*\\%
    % red,densely dashed,every mark/.append style={solid,fill=red!80!black},mark=*\\%
    % brown!60!black,densely dashed,every mark/.append style={
    % solid,fill=brown!80!black},mark=square*\\%
    % black,densely dashed,every mark/.append style={solid,fill=gray},mark=otimes*\\%
    % blue,densely dashed,mark=star,every mark/.append style=solid\\%
    % red,densely dashed,every mark/.append style={solid,fill=red!80!black},mark=diamond*\\%
    % }
    \begin{axis}[
    	xlabel={Inference Speed [images/sec]},
    	ylabel={Top-1 Accuracy [\%]},
    % 	cycle list name=mycolorlist,
    	width=\textwidth*0.9,
    	height=\axisdefaultwidth*0.87,
    legend style={anchor=north east},
    	legend entries={Transformer,
                CNN,
                Mobile,
                MLP-Only},
                grid=major,
    ]
    \addlegendentry[mark=square*, color=blue]{Transformer}
    \addlegendentry[mark=*, color=red]{CNN}
    \addlegendentry[mark=triangle*, color=orange]{Mobile-Oriented}
    \addlegendentry[mark=star*, color=darkgreen]{MLP-only}
    
    \addplot [mark=square*,  color=blue,
        visualization depends on=\thisrow{alignment} \as \alignment,
        nodes near coords, % Place nodes near each coordinate
        point meta=explicit symbolic, % The meta data used in the nodes is not explicitly provided and not numeric
        every node near coord/.style={anchor=\alignment} % Align each coordinate at the anchor 40 degrees clockwise from the right edge
        ] table [% Provide data as a table
         meta index=2 % the meta data is found in the third column
         ] {
    x       y       label       alignment
    60.3     84.2    \textbf{LeViT-768*}      190
    };
    
    \addplot [mark=square*,  color=blue,
        visualization depends on=\thisrow{alignment} \as \alignment,
        nodes near coords, % Place nodes near each coordinate
        point meta=explicit symbolic, % The meta data used in the nodes is not explicitly provided and not numeric
        every node near coord/.style={anchor=\alignment} % Align each coordinate at the anchor 40 degrees clockwise from the right edge
        ] table [% Provide data as a table
         meta index=2 % the meta data is found in the third column
         ] {
    x       y       label       alignment
    136.22     82.5    \textbf{LeViT-384}      180
    };

    \addplot [mark=asterisk, color=red,
        visualization depends on=\thisrow{alignment} \as \alignment,
        nodes near coords, % Place nodes near each coordinate
        point meta=explicit symbolic, % The meta data used in the nodes is not explicitly provided and not numeric
        every node near coord/.style={anchor=\alignment} % Align each coordinate at the anchor 40 degrees clockwise from the right edge
        ] table [% Provide data as a table
         meta index=2 % the meta data is found in the third column
         ] {
    x       y       label       alignment
    39.73    84    ConvNext-S      -30
    };
    \addplot [mark=triangle*, color=orange,
        visualization depends on=\thisrow{alignment} \as \alignment,
        nodes near coords, % Place nodes near each coordinate
        point meta=explicit symbolic, % The meta data used in the nodes is not explicitly provided and not numeric
        every node near coord/.style={anchor=\alignment} % Align each coordinate at the anchor 40 degrees clockwise from the right edge
        ] table [% Provide data as a table
         meta index=2 % the meta data is found in the third column
         ] {
    x       y       label       alignment
    498.32    80.6    OFA-595m      180
    };
    \addplot [mark=o, color=darkgreen,
        visualization depends on=\thisrow{alignment} \as \alignment,
        nodes near coords, % Place nodes near each coordinate
        point meta=explicit symbolic, % The meta data used in the nodes is not explicitly provided and not numeric
        every node near coord/.style={anchor=\alignment} % Align each coordinate at the anchor 40 degrees clockwise from the right edge
        ] table [% Provide data as a table
         meta index=2 % the meta data is found in the third column
         ] {
    x       y       label       alignment
    139.60    78.8    ResMLP-S12      180
    };
    \addplot [mark=square*, color=blue,
        visualization depends on=\thisrow{alignment} \as \alignment,
        nodes near coords, % Place nodes near each coordinate
        point meta=explicit symbolic, % The meta data used in the nodes is not explicitly provided and not numeric
        every node near coord/.style={anchor=\alignment} % Align each coordinate at the anchor 40 degrees clockwise from the right edge
        ] table [% Provide data as a table
         meta index=2 % the meta data is found in the third column
         ] {
    x       y       label       alignment
    34.78    84    Swin-S      -325
    };
    \addplot [mark=asterisk, color=red,
        visualization depends on=\thisrow{alignment} \as \alignment,
        nodes near coords, % Place nodes near each coordinate
        point meta=explicit symbolic, % The meta data used in the nodes is not explicitly provided and not numeric
        every node near coord/.style={anchor=\alignment} % Align each coordinate at the anchor 40 degrees clockwise from the right edge
        ] table [% Provide data as a table
         meta index=2 % the meta data is found in the third column
         ] {
    x       y       label       alignment
    88.54    82.4    TResNet-M      -200
    };
    \addplot [mark=asterisk, color=red,
        visualization depends on=\thisrow{alignment} \as \alignment,
        nodes near coords, % Place nodes near each coordinate
        point meta=explicit symbolic, % The meta data used in the nodes is not explicitly provided and not numeric
        every node near coord/.style={anchor=\alignment} % Align each coordinate at the anchor 40 degrees clockwise from the right edge
        ] table [% Provide data as a table
         meta index=2 % the meta data is found in the third column
         ] {
    x       y       label       alignment
    55.12    83.9    TResNet-L      170
    };
    \addplot [mark=square*, color=blue,
        visualization depends on=\thisrow{alignment} \as \alignment,
        nodes near coords, % Place nodes near each coordinate
        point meta=explicit symbolic, % The meta data used in the nodes is not explicitly provided and not numeric
        every node near coord/.style={anchor=\alignment} % Align each coordinate at the anchor 40 degrees clockwise from the right edge
        ] table [% Provide data as a table
         meta index=2 % the meta data is found in the third column
         ] {
    x       y       label       alignment
    76.66    81.8    ViT-S      -195
    };
    \addplot [mark=square*,  color=blue,
        visualization depends on=\thisrow{alignment} \as \alignment,
        nodes near coords, % Place nodes near each coordinate
        point meta=explicit symbolic, % The meta data used in the nodes is not explicitly provided and not numeric
        every node near coord/.style={anchor=\alignment} % Align each coordinate at the anchor 40 degrees clockwise from the right edge
        ] table [% Provide data as a table
         meta index=2 % the meta data is found in the third column
         ] {
    x       y       label       alignment
    263.03    81.5    \textbf{LeViT-256}      150
    };
    \addplot [mark=asterisk, color=red,
        visualization depends on=\thisrow{alignment} \as \alignment,
        nodes near coords, % Place nodes near each coordinate
        point meta=explicit symbolic, % The meta data used in the nodes is not explicitly provided and not numeric
        every node near coord/.style={anchor=\alignment} % Align each coordinate at the anchor 40 degrees clockwise from the right edge
        ] table [% Provide data as a table
         meta index=2 % the meta data is found in the third column
         ] {
    x       y       label       alignment
    117.08    80.9    ResNet-50      190
    };
    \addplot [mark=asterisk, color=red,
        visualization depends on=\thisrow{alignment} \as \alignment,
        nodes near coords, % Place nodes near each coordinate
        point meta=explicit symbolic, % The meta data used in the nodes is not explicitly provided and not numeric
        every node near coord/.style={anchor=\alignment} % Align each coordinate at the anchor 40 degrees clockwise from the right edge
        ] table [% Provide data as a table
         meta index=2 % the meta data is found in the third column
         ] {
    x       y       label       alignment
    107.91    82.8    RegNetY-32      -10
    };

    \addplot [mark=triangle*, color=orange,
        visualization depends on=\thisrow{alignment} \as \alignment,
        nodes near coords, % Place nodes near each coordinate
        point meta=explicit symbolic, % The meta data used in the nodes is not explicitly provided and not numeric
        every node near coord/.style={anchor=\alignment} % Align each coordinate at the anchor 40 degrees clockwise from the right edge
        ] table [% Provide data as a table
         meta index=2 % the meta data is found in the third column
         ] {
    x       y       label       alignment
    802.78    77.3    MobileNetV3-L      270
    };
    %     \addplot [mark=triangle*, color=orange,
    %     visualization depends on=\thisrow{alignment} \as \alignment,
    %     nodes near coords, % Place nodes near each coordinate
    %     point meta=explicit symbolic, % The meta data used in the nodes is not explicitly provided and not numeric
    %     every node near coord/.style={anchor=\alignment} % Align each coordinate at the anchor 40 degrees clockwise from the right edge
    %     ] table [% Provide data as a table
    %      meta index=2 % the meta data is found in the third column
    %      ] {
    % x       y       label       alignment
    % 1080    77.8    OFA-Pixel62ms      270
    % };
    \addplot [mark=triangle*, color=orange,
        visualization depends on=\thisrow{alignment} \as \alignment,
        nodes near coords, % Place nodes near each coordinate
        point meta=explicit symbolic, % The meta data used in the nodes is not explicitly provided and not numeric
        every node near coord/.style={anchor=\alignment} % Align each coordinate at the anchor 40 degrees clockwise from the right edge
        ] table [% Provide data as a table
         meta index=2 % the meta data is found in the third column
         ] {
    x       y       label       alignment
    561.55    79.3    EfficientNet-B0      180
    };
    \addplot [mark=triangle*, color=orange,
        visualization depends on=\thisrow{alignment} \as \alignment,
        nodes near coords, % Place nodes near each coordinate
        point meta=explicit symbolic, % The meta data used in the nodes is not explicitly provided and not numeric
        every node near coord/.style={anchor=\alignment} % Align each coordinate at the anchor 40 degrees clockwise from the right edge
        ] table [% Provide data as a table
         meta index=2 % the meta data is found in the third column
         ] {
    x       y       label       alignment
    248.7    82.1    EfficientNet-B3      180
    };
    \addplot [mark=triangle*, color=orange,
        visualization depends on=\thisrow{alignment} \as \alignment,
        nodes near coords, % Place nodes near each coordinate
        point meta=explicit symbolic, % The meta data used in the nodes is not explicitly provided and not numeric
        every node near coord/.style={anchor=\alignment} % Align each coordinate at the anchor 40 degrees clockwise from the right edge
        ] table [% Provide data as a table
         meta index=2 % the meta data is found in the third column
         ] {
    x       y       label       alignment
    358.94    81.29      EfficientNet-B2      180
    };
    \addplot [mark=square*, color=blue,
        visualization depends on=\thisrow{alignment} \as \alignment,
        nodes near coords, % Place nodes near each coordinate
        point meta=explicit symbolic, % The meta data used in the nodes is not explicitly provided and not numeric
        every node near coord/.style={anchor=\alignment} % Align each coordinate at the anchor 40 degrees clockwise from the right edge
        ] table [% Provide data as a table
         meta index=2 % the meta data is found in the third column
         ] {
    x       y       label       alignment
    42.82    82.5    TnT-S      1
    };
    \addplot [mark=o, color=darkgreen,
        visualization depends on=\thisrow{alignment} \as \alignment,
        nodes near coords, % Place nodes near each coordinate
        point meta=explicit symbolic, % The meta data used in the nodes is not explicitly provided and not numeric
        every node near coord/.style={anchor=\alignment} % Align each coordinate at the anchor 40 degrees clockwise from the right edge
        ] table [% Provide data as a table
         meta index=2 % the meta data is found in the third column
         ] {
    x       y       label       alignment
    32.5    82    Mixer-B      -270
    };

    % \legend{EfficientNet,TResNet}
    \end{axis}
    % \matrix [draw,below left] at (current bounding box.north east) {
    %   \node [blue,label=right:Transformer] {}; \\
    % };
   
    \end{tikzpicture}
    \caption{Speed-Accuracy comparison on an Intel Xeon Cascade Lake 2.5Ghz CPU, using OpenVINO inference engine.}
    \label{fig:speed_acc_cpu}
% \vspace{-0.2cm}    
\end{figure*}

\paragraph{Implementation details}
On GPU, The throughput was tested with TensorRT inference engine, FP16, and batch size of 256.
On CPU, The throughput was tested using Intel's OpenVINO Inference Engine,  FP16, a batch size of 1 and 16 streams (equivalent to the number of CPU cores).
All measurements were done after models were optimized to inference by batch-norm fusion. This significantly accelerates models that utilize batch-norm, like ResNet50, TResNet and LeViT. Note that LeViT-768* model is not a part of the original paper~\cite{graham2021levit}, but a model defined by us, to test LeViT design on higher accuracies regime.

\paragraph{GPU inference analysis}
On GPU, for low-to-medium accuracies, the most efficient models are LeViT-256 and LeViT-384. For higher accuracies ($>83.5\%$), TResNet-L and LeViT-768* models provides the best trade-off among the models tested.
Note that besides LeViT, other transformer models, such as Swin, TnT and ViT, provide inferior speed-accuracy trade-off compared to modern CNNs. Mobile-oriented models also do not excel on GPU inference, compared to top CNNs. 
Also note that several modern architectures, titled "Small" (ConvNext-S, Swin-S, TnT-S), are in fact quite resource-intensive - their inference speed is approximately three times slower compared to a plain ResNet50 model.

\paragraph{CPU inference analysis}
On CPU, Mobile oriented models (OFA, MobileNet, EfficientNet) provide the best speed-accuracy trade-off. LeViT models, who excelled on GPU inference, are not as efficient for CPU inference.

\subsection{Additional Ablations}
In this section we will present additional ablations and tests, to further validate our USI training configuration.
\vspace{-0.2cm}
\subsubsection{KD teacher relative weight}
In Table \ref{Table:kd_alpha} we test the impact of the relative weight between the KD loss and and supervision loss ($\alpha_{kd}$ in Eq. \ref{eq:KD_loss}).
\begin{table}[hbt!]
\centering

\begin{tabular}{c|c} 
\Xhline{3\arrayrulewidth}
KD relative weight, $\alpha_{kd}$ & Top1 Acc. [\%]  \\ 
\Xhline{3\arrayrulewidth}

0 (no KD loss)           & 76.2             \\
1      & 80.8             \\
5       & 82.7   \\
10       & 82.6   \\
20       & 82.7   \\
$\infty$ (no CE loss)     & 82.7   \\
\hline
\end{tabular}
\caption{Accuracy for different KD relative weights. Model tested - LeViT-384}
\label{Table:kd_alpha}
\vspace{-0.2cm}
\end{table}

As can be seen, without KD ($\alpha_{kd}=0$), our training scheme performs  poorly - $6.5\%$ less than accuracy obtained with the default value, $\alpha_{kd}=5$. If the KD relative weight is too low ($\alpha_{kd}=1$), there is also a decline in scores. For $\alpha_{kd}\geq
5$ we achieve top results. Interestingly, results remain the same even when training without the original hard-label supervision, and relying  only on the teacher ($\alpha_{kd}=\infty)$. This further demonstrates the effectiveness of KD in ImageNet training.

\subsubsection{KD Temperature}
In Table \ref{Table:kd_temperature} we investigate the impact of KD Temperature ($\tau$ in Eq. \ref{eq:KD_loss}) on the accuracy.
\begin{table}[hbt!]
\centering

\begin{tabular}{c|c} 
\Xhline{3\arrayrulewidth}
KD Temperature, $\tau$ & Top1 Acc. [\%]  \\ 
\Xhline{3\arrayrulewidth}
0.1   & 79.3             \\
1         & 82.7             \\
2       & 82.7             \\
5       & 81.7   \\
10       & 81.4   \\
\hline
\end{tabular}
\caption{Accuracy for different KD temperatures. Model tested - LeViT-384}
\label{Table:kd_temperature}
\vspace{-0.2cm}
\end{table}

Table \ref{Table:kd_temperature} shows that there is no benefit from using temperature in our KD loss. Both $\tau<1$ (sharpening the teacher predictions) and  $\tau>1$ (softening the teacher predictions) reduce the accuracy.
Utilizing vanilla softmax probabilities leads to the best results.

\subsubsection{Mixup-Cutmix vs. Cutout}
Cutout, Mixup and Cutmix are important  augmentations, that  significantly improved ImageNet Scores~\cite{ridnik2021tresnet}.
While Cutout only augments the input image, Cutmix and Mixup also alter the labels.
Cutout is more prevalent when training CNNs and Mobile-oriented models~\cite{howard2019searching,he2019bag}, while a combination of Mixup and Cutmix, as proposed in~\cite{rw2019timm}, is popular for training Transformer-based models~\cite{graham2021levit,touvron2021training}.
In Table \ref{Table:mixup_vs_cutout} we compare using Cutout to Mixup-Cutmix in our unified scheme.
\begin{table}[hbt!]
\centering

\begin{tabular}{c|c} 
\Xhline{3\arrayrulewidth}
Augmentation Type & Top1 Acc. [\%]  \\ 
\Xhline{3\arrayrulewidth}
None          & 82.0             \\
Cutout          & 82.4             \\
Mixup-Cutmix       & 82.7   \\
\hline
\end{tabular}
\caption{Accuracy for different augmentations. Model tested - LeViT-384}
\label{Table:mixup_vs_cutout}
\end{table}

As can be seen, applying each augmentation  is beneficial, but Mixup-Cutmix augmentation contributes more to improving the accuracy.
\vspace{-0.2cm}
\subsubsection{Architecture-based regularizations}
It is common to apply architecture-based regularizations, such as drop-path~\cite{cai2019effective} and drop-block~\cite{ghiasi2018dropblock}, mainly when training Transformer-based models~\cite{dosovitskiy2020image,touvron2021training}. These regularizations are not always applicable to other types of architectures, and their implementations sometimes differ between different models. Hence we preferred to avoid using them in our unified scheme. 

In Table \ref{Table:drop_path} we test whether adding drop-path to our scheme, when training a Transformer-based model, would improve results.

\begin{table}[hbt!]
\centering

\begin{tabular}{c|c} 
\Xhline{3\arrayrulewidth}
Drop-path & Top1 Acc. [\%]  \\ 
\Xhline{3\arrayrulewidth}

0         & 82.7             \\
0.1       & 82.6             \\
0.2       & 82.5   \\
\hline
\end{tabular}
\caption{Accuracy for different values of drop-path regularization. Model tested - LeViT-384}
\label{Table:drop_path}
\vspace{-0.2cm}
\end{table}

As can be seen, there is no gain from adding drop-path regularizations to \method{} scheme.

\subsubsection{Repository}
Due to implementation details and other factors, training schemes on a specific repository sometimes under-perform when used by a different repository.
We developed \method{} on an inner private repository. To increase our confidence in its validity, we re-implemented it on a publicly available repository - timm~\cite{rw2019timm}. We were able to reproduce on timm all the results obtained on our private repo. In \footnote{\url{https://github.com/Alibaba-MIIL/Solving\_ImageNet}} we will share our timm-based implementation.

\section{Discussion and Conclusions}

In this paper, we introduced a unified scheme for ImageNet training, called \method{}. \method{}, which utilizes KD and modern training tricks, requires no hyper-parameters tuning between different models, and enables to train any backbone to top results. Hence, it transforms training on ImageNet from an expert-oriented task to an automatic procedure.
With \method{}, we are also able to perform a methodical speed-accuracy comparison, and reliably identify efficient computer-vision backbones.

\paragraph{Applicability to other classification datasets}
For ImageNet, we are training a model from scratch. This implies using high learning rates, strong regularization, and more epochs. Hence, our ImageNet-dedicated \method{} scheme is not directly applicable to other classification datasets, which usually use \textit{transfer learning}. However, KD remains a highly effective technique also for the transfer learning case. 
It not only contributes to improving scores, but also leads to a more robust training procedure,  which is less sensitive to hyper-parameter tuning. We believe that various AutoML schemes would benefit from KD usage, and plan to demonstrate this in future work.

{\small
\bibliographystyle{ieee_fullname.bst}
\bibliography{egbib.bib}
}

\clearpage

\onecolumn

\appendix

\begin{appendices}

\section{Model Details}

In Table \ref{Table:model_details} we provide full details for various models. Measurements were done using PyTorch engine.

\begin{table}[hbt!]
\centering
\begin{tabular}{c|c|c|c|c}
\Xhline{3\arrayrulewidth}

Model name       & \begin{tabular}[c]{@{}c@{}}Top-1 Accuracy \\{[}\%]\end{tabular} & \begin{tabular}[c]{@{}c@{}}Maximal Inference Speed\\{[}img/sec]\end{tabular} & \begin{tabular}[c]{@{}c@{}}Maximal Training Speed\\{[}img/sec]\end{tabular} & \begin{tabular}[c]{@{}c@{}}Maximal \\Batch Size\end{tabular}  \\ 
\Xhline{3\arrayrulewidth}
MobileNetV3      & 77.3                                                            & 6359                                                                         & 1316                                                                        & 480                                                           \\
ResMLP-S12       & 78.8                                                            & 4909                                                                         & 1474                                                                        & 440                                                           \\
EfficienetNet-B0 & 79.3                                                            & 5112                                                                         & 1047                                                                        & 316                                                           \\
OFA-595m         & 80.6                                                            & 3919                                                                         & 601                                                                         & 292                                                           \\
ResNet50         & 81.0                                                            & 2819                                                                         & 785                                                                         & 316                                                           \\
ViT-S            & 81.8                                                            & 2432                                                                         & 724                                                                         & 244                                                           \\
Mixer-B          & 82.0                                                            & 1388                                                                         & 457                                                                         & 168                                                           \\
EfficienetNet-B3 & 82.1                                                            & 2604                                                                         & 511                                                                         & 168                                                           \\
TResNet-M        & 82.4                                                            & 2917                                                                         & 720                                                                         & 504                                                           \\
TnT-S            & 82.6                                                            & 782                                                                          & 152                                                                         & 146    \\
LeViT-384        & 82.7                                                            & 5175                                                                         & 1400                                                                        & 448                                                           \\
LeViT-768        & 84.2                                                            & 2116                                                                         & 594                                                                        & 196                                                           \\
RegNetY-32       & 82.8                                                            & 2013                                                                         & 546                                                                         & 260                                                           \\

TResNet-L        & 83.9                                                            & 1563                                                                         & 340                                                                         & 228                                                           \\

Swin-S           & 84.0                                                            & 858                                                                          & 223                                                                         & 112                                                           \\
ConvNext-S       & 84.0                                                            & 1172                                                                         & 350                                                                         & 128                                                           \\
\hline                                                      
\end{tabular}
\caption{Full model details. All measurements were done on Nvidia V100 GPU, with 16GB. Train and test resolution - $224$.}
\label{Table:model_details}
\end{table}

\section{GPU Inference using PyTorch inference engine}
\pgfplotsset{grid style={dashed}}
\begin{figure*}
    \centering
    \begin{tikzpicture}
    \definecolor{darkgreen}{rgb}{0.0, 0.5, 0.0}
    % \pgfplotscreateplotcyclelist{mycolorlist}{%
    % blue,every mark/.append style={fill=blue!80!black},mark=*\\%
    % red,every mark/.append style={fill=red!80!black},mark=square*\\%
    % brown!60!black,every mark/.append style={fill=brown!80!black},mark=otimes*\\%
    % black,mark=star\\%
    % blue,every mark/.append style={fill=blue!80!black},mark=diamond*\\%
    % red,densely dashed,every mark/.append style={solid,fill=red!80!black},mark=*\\%
    % brown!60!black,densely dashed,every mark/.append style={
    % solid,fill=brown!80!black},mark=square*\\%
    % black,densely dashed,every mark/.append style={solid,fill=gray},mark=otimes*\\%
    % blue,densely dashed,mark=star,every mark/.append style=solid\\%
    % red,densely dashed,every mark/.append style={solid,fill=red!80!black},mark=diamond*\\%
    % }
    \begin{axis}[
    	xlabel={Inference Speed [images/sec]},
    	ylabel={Top-1 Accuracy [\%]},
    % 	cycle list name=mycolorlist,
    	width=\textwidth*0.9,
    	height=\axisdefaultwidth*0.88,
        legend style={anchor=north east},
    	legend entries={Transformer,
                CNN,
                Mobile,
                MLP-Only},
        xmin=200,
        xmax=8200,  
        grid=major,
        xtick={1000,2000,3000,4000,5000,6000,7000,8000},
        xticklabels={1000,2000,3000,4000,5000,6000,7000,8000}
    ]
    \addlegendentry[mark=square*, color=blue]{Transformer}
    \addlegendentry[mark=*, color=red]{CNN}
    \addlegendentry[mark=triangle*, color=orange]{Mobile-Oriented}
    \addlegendentry[mark=star*, color=darkgreen]{MLP-only}
    
    \addplot [mark=square*, color=blue,
        visualization depends on=\thisrow{alignment} \as \alignment,
        nodes near coords, % Place nodes near each coordinate
        point meta=explicit symbolic, % The meta data used in the nodes is not explicitly provided and not numeric
        every node near coord/.style={anchor=\alignment} % Align each coordinate at the anchor 40 degrees clockwise from the right edge
        ] table [% Provide data as a table
         meta index=2 % the meta data is found in the third column
         ] {
    x       y       label       alignment
    2166    84.2    \textbf{LeViT-768*}     180
    };

    \addplot [mark=square*,  color=blue,
        visualization depends on=\thisrow{alignment} \as \alignment,
        nodes near coords, % Place nodes near each coordinate
        point meta=explicit symbolic, % The meta data used in the nodes is not explicitly provided and not numeric
        every node near coord/.style={anchor=\alignment} % Align each coordinate at the anchor 40 degrees clockwise from the right edge
        ] table [% Provide data as a table
         meta index=2 % the meta data is found in the third column
         ] {
    x       y       label       alignment
    5175    82.7    \textbf{LeViT-384}      -90
    };

    \addplot [mark=asterisk, color=red,
        visualization depends on=\thisrow{alignment} \as \alignment,
        nodes near coords, % Place nodes near each coordinate
        point meta=explicit symbolic, % The meta data used in the nodes is not explicitly provided and not numeric
        every node near coord/.style={anchor=\alignment} % Align each coordinate at the anchor 40 degrees clockwise from the right edge
        ] table [% Provide data as a table
         meta index=2 % the meta data is found in the third column
         ] {
    x       y       label       alignment
    1172    84    ConvNext-S      220
    };
    \addplot [mark=triangle*, color=orange,
        visualization depends on=\thisrow{alignment} \as \alignment,
        nodes near coords, % Place nodes near each coordinate
        point meta=explicit symbolic, % The meta data used in the nodes is not explicitly provided and not numeric
        every node near coord/.style={anchor=\alignment} % Align each coordinate at the anchor 40 degrees clockwise from the right edge
        ] table [% Provide data as a table
         meta index=2 % the meta data is found in the third column
         ] {
    x       y       label       alignment
    3919    80.6    OFA-595m      180
    };
    \addplot [mark=o, color=darkgreen,
        visualization depends on=\thisrow{alignment} \as \alignment,
        nodes near coords, % Place nodes near each coordinate
        point meta=explicit symbolic, % The meta data used in the nodes is not explicitly provided and not numeric
        every node near coord/.style={anchor=\alignment} % Align each coordinate at the anchor 40 degrees clockwise from the right edge
        ] table [% Provide data as a table
         meta index=2 % the meta data is found in the third column
         ] {
    x       y       label       alignment
    4909.1    78.8    ResMLP-S12      180
    };
    \addplot [mark=square*, color=blue,
        visualization depends on=\thisrow{alignment} \as \alignment,
        nodes near coords, % Place nodes near each coordinate
        point meta=explicit symbolic, % The meta data used in the nodes is not explicitly provided and not numeric
        every node near coord/.style={anchor=\alignment} % Align each coordinate at the anchor 40 degrees clockwise from the right edge
        ] table [% Provide data as a table
         meta index=2 % the meta data is found in the third column
         ] {
    x       y       label       alignment
    840    84    Swin-S      -270
    };
    \addplot [mark=asterisk, color=red,
        visualization depends on=\thisrow{alignment} \as \alignment,
        nodes near coords, % Place nodes near each coordinate
        point meta=explicit symbolic, % The meta data used in the nodes is not explicitly provided and not numeric
        every node near coord/.style={anchor=\alignment} % Align each coordinate at the anchor 40 degrees clockwise from the right edge
        ] table [% Provide data as a table
         meta index=2 % the meta data is found in the third column
         ] {
    x       y       label       alignment
    2917    82.4    TResNet-M      -90
    };
    \addplot [mark=asterisk, color=red,
        visualization depends on=\thisrow{alignment} \as \alignment,
        nodes near coords, % Place nodes near each coordinate
        point meta=explicit symbolic, % The meta data used in the nodes is not explicitly provided and not numeric
        every node near coord/.style={anchor=\alignment} % Align each coordinate at the anchor 40 degrees clockwise from the right edge
        ] table [% Provide data as a table
         meta index=2 % the meta data is found in the third column
         ] {
    x       y       label       alignment
    1560    83.9    TResNet-L      180
    };
    \addplot [mark=square*, color=blue,
        visualization depends on=\thisrow{alignment} \as \alignment,
        nodes near coords, % Place nodes near each coordinate
        point meta=explicit symbolic, % The meta data used in the nodes is not explicitly provided and not numeric
        every node near coord/.style={anchor=\alignment} % Align each coordinate at the anchor 40 degrees clockwise from the right edge
        ] table [% Provide data as a table
         meta index=2 % the meta data is found in the third column
         ] {
    x       y       label       alignment
    2432    81.8    ViT-S      90
    };
    \addplot [mark=square*,  color=blue,
        visualization depends on=\thisrow{alignment} \as \alignment,
        nodes near coords, % Place nodes near each coordinate
        point meta=explicit symbolic, % The meta data used in the nodes is not explicitly provided and not numeric
        every node near coord/.style={anchor=\alignment} % Align each coordinate at the anchor 40 degrees clockwise from the right edge
        ] table [% Provide data as a table
         meta index=2 % the meta data is found in the third column
         ] {
    x       y       label       alignment
    7551    81.5    \textbf{LeViT-256}      0
    };
    \addplot [mark=asterisk, color=red,
        visualization depends on=\thisrow{alignment} \as \alignment,
        nodes near coords, % Place nodes near each coordinate
        point meta=explicit symbolic, % The meta data used in the nodes is not explicitly provided and not numeric
        every node near coord/.style={anchor=\alignment} % Align each coordinate at the anchor 40 degrees clockwise from the right edge
        ] table [% Provide data as a table
         meta index=2 % the meta data is found in the third column
         ] {
    x       y       label       alignment
    2819    80.9    ResNet-50      190
    };
    \addplot [mark=asterisk, color=red,
        visualization depends on=\thisrow{alignment} \as \alignment,
        nodes near coords, % Place nodes near each coordinate
        point meta=explicit symbolic, % The meta data used in the nodes is not explicitly provided and not numeric
        every node near coord/.style={anchor=\alignment} % Align each coordinate at the anchor 40 degrees clockwise from the right edge
        ] table [% Provide data as a table
         meta index=2 % the meta data is found in the third column
         ] {
    x       y       label       alignment
    1910    82.8    RegNetY-32      290
    };

    \addplot [mark=triangle*, color=orange,
        visualization depends on=\thisrow{alignment} \as \alignment,
        nodes near coords, % Place nodes near each coordinate
        point meta=explicit symbolic, % The meta data used in the nodes is not explicitly provided and not numeric
        every node near coord/.style={anchor=\alignment} % Align each coordinate at the anchor 40 degrees clockwise from the right edge
        ] table [% Provide data as a table
         meta index=2 % the meta data is found in the third column
         ] {
    x       y       label       alignment
    6359    77.3    MobileNetV3-L      180
    };
    \addplot [mark=triangle*, color=orange,
        visualization depends on=\thisrow{alignment} \as \alignment,
        nodes near coords, % Place nodes near each coordinate
        point meta=explicit symbolic, % The meta data used in the nodes is not explicitly provided and not numeric
        every node near coord/.style={anchor=\alignment} % Align each coordinate at the anchor 40 degrees clockwise from the right edge
        ] table [% Provide data as a table
         meta index=2 % the meta data is found in the third column
         ] {
    x       y       label       alignment
    4513    79.3    EfficientNet-B0      180
    };
    \addplot [mark=square*, color=blue,
        visualization depends on=\thisrow{alignment} \as \alignment,
        nodes near coords, % Place nodes near each coordinate
        point meta=explicit symbolic, % The meta data used in the nodes is not explicitly provided and not numeric
        every node near coord/.style={anchor=\alignment} % Align each coordinate at the anchor 40 degrees clockwise from the right edge
        ] table [% Provide data as a table
         meta index=2 % the meta data is found in the third column
         ] {
    x       y       label       alignment
    783    82.5    TnT-S      180
    };
    \addplot [mark=o, color=darkgreen,
        visualization depends on=\thisrow{alignment} \as \alignment,
        nodes near coords, % Place nodes near each coordinate
        point meta=explicit symbolic, % The meta data used in the nodes is not explicitly provided and not numeric
        every node near coord/.style={anchor=\alignment} % Align each coordinate at the anchor 40 degrees clockwise from the right edge
        ] table [% Provide data as a table
         meta index=2 % the meta data is found in the third column
         ] {
    x       y       label       alignment
    1388    82    Mixer-B      -290
    };
    
    \addplot [mark=triangle*, color=orange,
        visualization depends on=\thisrow{alignment} \as \alignment,
        nodes near coords, % Place nodes near each coordinate
        point meta=explicit symbolic, % The meta data used in the nodes is not explicitly provided and not numeric
        every node near coord/.style={anchor=\alignment} % Align each coordinate at the anchor 40 degrees clockwise from the right edge
        ] table [% Provide data as a table
         meta index=2 % the meta data is found in the third column
         ] {
    x       y       label       alignment
    2604    82.1      EfficientNet-B3      178
    };

    % \legend{EfficientNet,TResNet}
    \end{axis}
    % \matrix [draw,below left] at (current bounding box.north east) {
    %   \node [blue,label=right:Transformer] {}; \\
    % };
   
    \end{tikzpicture}
    \caption{Speed-Accuracy comparison on an Nvidia V100 GPU, with PyTorch inference engine.}
    \label{fig:speed_acc_v100}
\end{figure*}
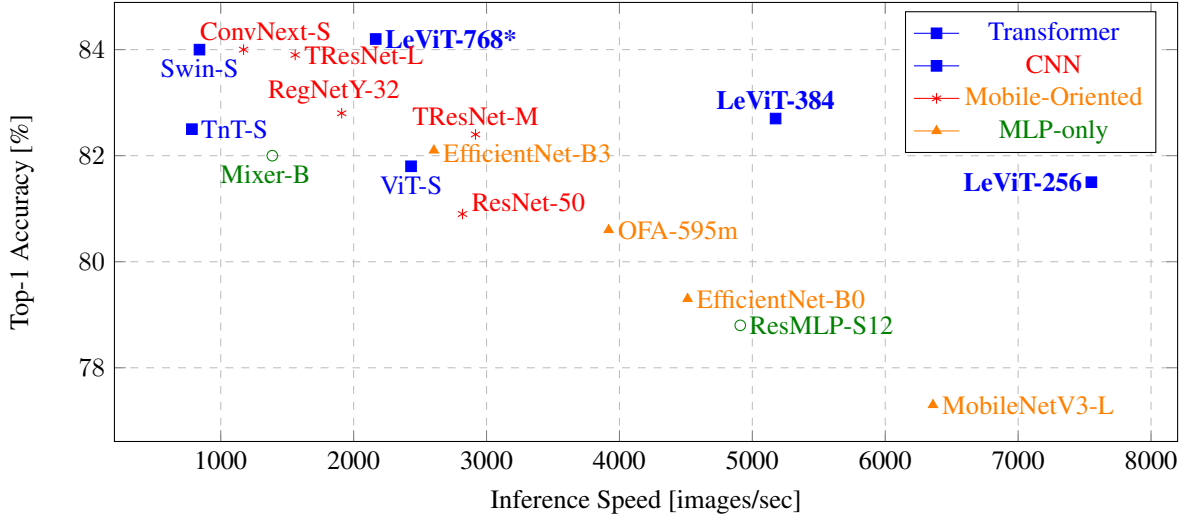
In Figure \ref{fig:speed_acc_v100} we compare GPU speed-accuracy trade off of different models with a PyTorch inference engine.

\end{appendices}

\end{document}